\documentclass[conference]{IEEEtran}
\usepackage{cite}
\usepackage{graphicx}
\usepackage{amsmath,amssymb,amsfonts}
\usepackage{algorithmic}
\usepackage{textcomp}
\usepackage{xcolor}

\usepackage{url}
\usepackage{tikz}
\usetikzlibrary{shapes.geometric, arrows, positioning}

\tikzstyle{block} = [rectangle, rounded corners, minimum width=3cm, minimum height=1cm, text centered, draw=black, fill=blue!10]
\tikzstyle{arrow} = [thick,->,>=stealth]
\tikzstyle{human} = [ellipse, minimum width=2cm, minimum height=1cm, text centered, draw=black, fill=red!10]
\tikzstyle{rdfnode} = [rectangle, draw=black, fill=green!10, text centered, minimum height=0.8cm]
\usepackage{forest}

\usepackage{hyperref}
\usepackage{cleveref}

\title{Ethical AI: Towards Defining a Collective Evaluation Framework}
\author{\IEEEauthorblockN{1\textsuperscript{st} Aasish Kumar Sharma}
\IEEEauthorblockA{\textit{Department of Computer Science} \\
\textit{Göttingent University}\\
Göttingen, Germany \\
aasish-kumar.sharma@gwdg.de}
\and
\IEEEauthorblockN{2\textsuperscript{nd} Dimitar Kyosev}
\IEEEauthorblockA{\textit{Department of Legal Affairs} \\
\textit{Alis Grave Nil Private Limited}\\
Burgas, Bulgaria \\
kyosev.dimitar@gmail.com}
\and
\IEEEauthorblockN{3\textsuperscript{th} Julian Kunkel}
\IEEEauthorblockA{\textit{Faculty of Mathematics and Computer Science} \\
\textit{Georg-August-Universität Göttingen}\\
Göttingen, Germany \\
julian.kunkel@gwdg.de}

}

\begin{document}

\maketitle

\section*{Notice}
This work has been accepted for presentation at the 8th IEEE International Workshop on Advances in Artificial Intelligence and Machine Learning (AIML 2025): Futuristic AI and ML models \& Intelligent Systems. © 2025 IEEE. Personal use of this material is permitted. The final published version will be available via IEEE Xplore at: \url{https://ieeexplore.ieee.org/}
\vspace{1em}

\begin{abstract}
Artificial Intelligence (AI) is transforming sectors such as healthcare, finance, and autonomous systems, offering powerful tools for innovation. Yet its rapid integration raises urgent ethical concerns related to data ownership, privacy, and systemic bias. Issues like opaque decision-making, misleading outputs, and unfair treatment in high-stakes domains underscore the need for transparent and accountable AI systems.

This article addresses these challenges by proposing a modular ethical assessment framework built on \textit{ontological blocks of meaning}—discrete, interpretable units that encode ethical principles such as fairness, accountability, and ownership. By integrating these blocks with FAIR (Findable, Accessible, Interoperable, Reusable) principles, the framework supports scalable, transparent, and legally aligned ethical evaluations, including compliance with the EU AI Act.

Using a real-world use case in AI-powered investor profiling, the paper demonstrates how the framework enables dynamic, behavior-informed risk classification. The findings suggest that ontological blocks offer a promising path toward explainable and auditable AI ethics, though challenges remain in automation and probabilistic reasoning.

\end{abstract}

\begin{IEEEkeywords}
Responsible AI, Ethical AI, Machine Ethics, AI Acts, Explainability, Ontology, Workflows, Transparency.
\end{IEEEkeywords}

\section{Introduction}
\label{sec:Introduction}
Artificial Intelligence (AI) has emerged as a transformative technology, revolutionizing fields such as healthcare, finance, and autonomous systems. While AI offers significant benefits, including enhanced productivity and innovative solutions to global challenges, its rapid evolution also introduces profound ethical challenges that require immediate and thorough attention.

One of the foundational challenges lies in the data used to train AI systems. AI relies heavily on vast datasets sourced from the internet, private organizations, and other repositories. However, issues related to data ownership, consent, and privacy remain unresolved. Questions such as "Who owns the data?" and "Was it ethically obtained?" highlight potential risks of misuse and exploitation, raising concerns about intellectual property rights and individual privacy.

Another pressing issue is trustworthiness. AI systems occasionally produce misleading or incorrect outputs, a phenomenon known as "hallucinations." Moreover, ensuring the safety of AI systems and maintaining detailed records of their decision-making processes are critical for fostering trust. Without robust mechanisms for safety and accountability, reliance on AI systems could lead to unforeseen risks, particularly in high-stakes domains such as healthcare, finance, and law enforcement.

However, the most profound challenges arise from the ethical implications of AI’s outcomes. AI systems, often operating as "black boxes", generate decisions and recommendations with far-reaching consequences. The absence of embedded ethical considerations can result in biased, unfair, or even harmful outcomes. For example:
\begin{itemize}
\item Biased algorithms in hiring or lending processes can perpetuate societal inequalities.
\item Unethical use of AI in surveillance can infringe on civil liberties.
\item Lack of explainability can erode trust, particularly in life-critical applications.
\end{itemize}

Addressing these challenges requires a collaborative effort among technologists, policymakers, ethicists, and the public. Establishing robust frameworks for ethical AI development and deployment is imperative to ensure AI aligns with human values and serves humanity equitably and responsibly.

\subsection{Key Ethical Challenges in AI}
\begin{itemize}
\item \textit{Data-Related Issues:} These include questions around data accuracy, inclusion of purposefully misleading data or even data poisoning 
\item \textit{Trustworthiness:} AI systems must be reliable, safe, and explainable to avoid risks like hallucinations and over reliance on black-box outcomes.
\item \textit{Ethical Implications of Outcomes:} The potential for biased or harmful decisions underscores the need to embed ethical principles directly into AI systems.
\end{itemize}

As AI continues to integrate into society, the task of addressing these ethical dilemmas grows increasingly complex. While the challenges are significant, they also present an opportunity to build a future where AI aligns with the highest ethical standards. By fostering collaboration and establishing comprehensive frameworks, we can ensure that AI development progresses in a way that prioritizes transparency, fairness, and accountability.

The article is organized in five sections: Introduction (\Cref{sec:Introduction}), background (\Cref{sec:Background}), literature review (\Cref{sec:LiteratureReview}), methodology and discussion (\Cref{sec:MethodologyAndDiscussion}), then includes conclusion and recommendations (\Cref{sec:ConclusionAndRecommendation}).

\section{Background}
\label{sec:Background}
Ethical AI refers to the development and deployment of artificial intelligence systems that emphasize fairness, transparency, accountability, and respect for fundamental rights \cite{jobin2019principles}. AI ethics emphasises the impact of AI on individuals, groups and wider society. The goal is to promote safe and responsible AI use, to mitigate AI’s novel risks and prevent harm. Much of the work in this area centres around four main verticals:
Frameworks such as the six core principles of the WHO\cite{abujaber2024ethical}, the EU AI Act\cite{EUAIAct_regulation_2024} highlight the importance of protecting autonomy, promoting equity, and fostering responsibility. Despite these efforts, ethical frameworks often lack technical grounding and do not address the dynamic nature of AI systems \cite{mika2019ethical, leslie2019understanding, ethics_of_ai, brey2024ethics, hanssen2019ethics, jobin2019principles}.

For example, Reinforcement-Learning (RL), particularly deep reinforcement learning, operates as a closed-box system, posing challenges in explainability\cite{linardatos_explainable_2021, vasan2024revisiting, ComprehensiveOverviewReward}. Human-in-the-loop (HITL) AI approaches\cite{mosqueira2023human} provide a potential solution by integrating human oversight at critical stages, but their implementation in Self Reinforcement Learning (SLR) remains limited.

Recent studies highlight several state-of-the-art advances in ethical AI and SRL. Mehrabi et al.\cite{mehrabi2021bias} explore bias mitigation techniques such as data cleaning, model adjustments, and output tuning. \cite{prem2023ethical, salih2024perspective} discusses interpretability approaches such as SHAP and LIME, which enhance explainability in complex models. Human-in-the-loop systems\cite{mosqueira2023human} integrate ethical oversight effectively, while the EU AI Act\cite{EUAIAct_regulation_2024} offers risk-based regulations tailored for high-stakes applications.

\subsection{Objectives}

This article addresses key challenges in ethical AI and aims to:

\begin{enumerate}
    \item Propose a unified system for monitoring AI outcomes in line with the EU AI Act and emerging global regulations.
    \item Evaluate the feasibility of using ontological blocks of meaning for ethical assessment.
    \item Explore the integration of FAIR (Findable, Accessible, Interoperable, Reusable) principles into these ontological blocks.
\end{enumerate}

Together, these objectives support the development of ethical AI frameworks that promote transparency, accountability, and fairness \cite{rossi2018ethically}.

\section{Literature Review}
\label{sec:LiteratureReview}

The rising importance of ethical Artificial Intelligence (AI) has driven researchers and institutions to develop frameworks, methods, and applications that promote transparency, accountability, and fairness. This review is organized into subsections covering finance, healthcare, autonomous vehicles, ontological and FAIR approaches, and global initiatives.

\subsection{Ethical AI in Finance}

AI-driven credit scoring in finance raises ethical concerns about bias and unequal treatment.

\begin{itemize}
    \item Hassani \cite{hassani2021bias} shows how societal biases in training data reinforce discrimination in credit assessments, disproportionately affecting marginalized groups.
    \item Packin \cite{packin2021disability} warns that lack of comprehensive data on disabled individuals can lead to exclusionary outcomes in AI-based scoring.
\end{itemize}

\textbf{Identified Gap:} Independent tools like ontological blocks are proposed to detect and mitigate embedded bias in financial AI systems.

\subsection{Ethical AI in Healthcare}

Healthcare—particularly oncology—illustrates the need for ethical AI frameworks to safeguard privacy, transparency, and fairness \cite{talati2023ai}.

\begin{itemize}
    \item Ontological blocks have been applied to sensitive patient data under FAIR principles, aiding in personalized treatment planning \cite{kumar2005oncology, lin2018cctoo}.
    \item Monteith et al.\ \cite{monteith2024misinformation} caution against misinformation in AI mental health tools, stressing the need for explainability and oversight.
    \item Antoniou et al.\ \cite{antoniou2022explainable} highlight the need for explainable AI to manage complex diagnoses involving comorbidities.
\end{itemize}

\textbf{Identified Gap:} Real-time, scalable use of explainable AI methods (e.g., SHAP, LIME) remains a key challenge in high-stakes clinical settings \cite{linardatos_explainable_2021}.

\subsection{Ethical AI in Autonomous Vehicles}

Autonomous vehicles demand AI systems that prioritize safety, accountability, and ethical reasoning.

\begin{itemize}
    \item Lin \cite{lin2016ethics} and Jenkins et. al. \cite{jenkins2022autonomous} examines scenarios like the trolley problem, calling for transparent ethical decision-making in critical moments.
    \item Goodall \cite{goodall2014ethics} advocates for programming rules that minimize harm during unavoidable accidents.
    \item Nyholm and Smids \cite{nyholm2016autonomous} argue for clear accountability frameworks among manufacturers, users, and regulators.
\end{itemize}

\textbf{Identified Gap:} Ensuring real-time ethical decisions and system adaptability remains a core technical and ethical challenge.

\subsection{Ontological and FAIR Approaches for Ethical AI}

Ontological models and FAIR principles are emerging as foundational tools for ethical AI design.

\begin{itemize}
    \item Ontological blocks, drawing from Semantic Web standards \cite{mcguinness2004owl, antoniou2009web}, provide scalable representations of principles like non-maleficence and equity \cite{abujaber2024ethical}.
    \item FAIR principles \cite{brey2024ethics}—Findable, Accessible, Interoperable, and Reusable—support the openness and standardization of ethical modules.
    \item Guizzardi et al.\ \cite{guizzardi2023OntologyBased} emphasize the need for ontologies that model user-centric concerns like privacy, fairness, and risk.
\end{itemize}

\textbf{Challenges:} Implementing FAIR-aligned ontological blocks is constrained by the manual effort needed to build and maintain interoperable frameworks.

\subsection{Global and Collaborative Efforts}

International collaborations aim to mitigate societal-scale risks and build governance frameworks for ethical AI \cite{roberts2022meta, benjamins2020societal, hanssen2019ethics}.

\begin{itemize}
    \item The EU champions “ethics-by-design,” embedding ethics at the development stage \cite{brey2024ethics}.
    \item An international consortium proposes benchmarking societal risks, promoting transparency and cooperative oversight \cite{gruetzemacher2023International}.
    \item Responsible Research and Innovation (RRI) offers a governance model that aligns AI with societal values \cite{rri2023governance, field2024exploring, ivanova2023frames, volker2024translations, pacifico2018introducing}.
\end{itemize}

\textbf{Future Direction:} These efforts underscore the importance of global standards and shared accountability mechanisms for ethical AI deployment.

\section{Methodology and Discussion}
\label{sec:MethodologyAndDiscussion}

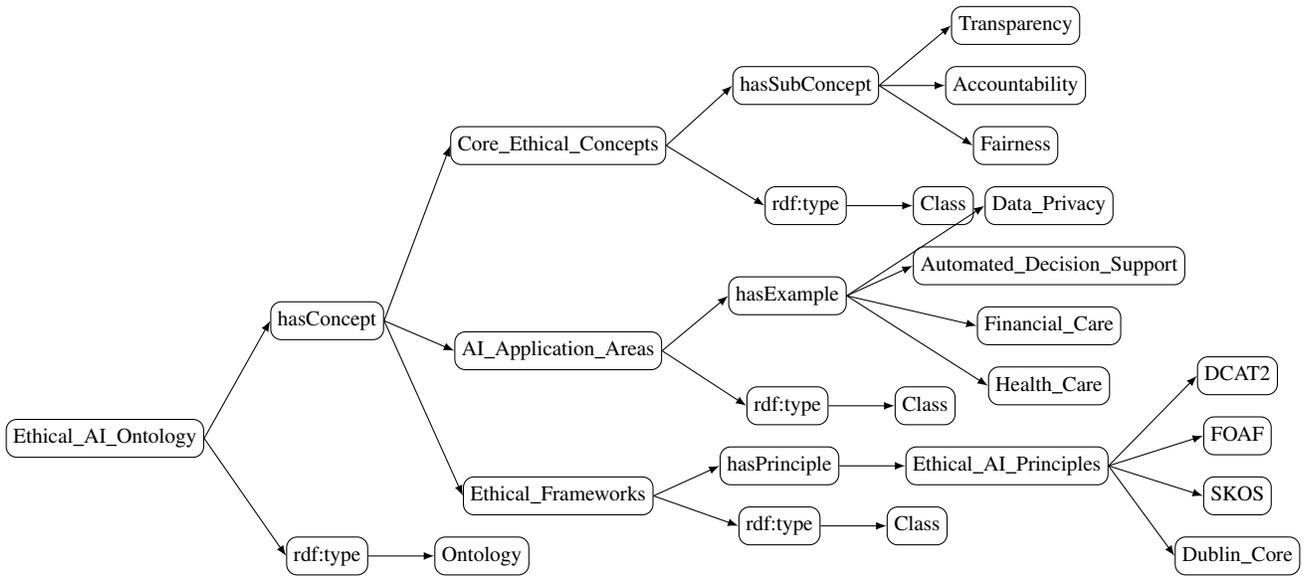
\begin{figure*}[ht!]
  \centering
  \footnotesize
  \begin{forest}
  for tree={
    grow=east,
    parent anchor=east,
    child anchor=west,
    align=left,
    edge={-latex},
    l sep=25pt,
    s sep=8pt,
    rounded corners,
    draw
  }
  [Ethical\_AI\_Ontology
    [rdf:type
      [Ontology]
    ]
    [hasConcept
      [Ethical\_Frameworks
        [rdf:type
          [Class]
        ]
        [hasPrinciple
          [Ethical\_AI\_Principles
            [Dublin\_Core]
            [SKOS]
            [FOAF]
            [DCAT2]
          ]
        ]
      ]
      [AI\_Application\_Areas
        [rdf:type
          [Class]
        ]
        [hasExample
          [Health\_Care]
          [Financial\_Care]
          [Automated\_Decision\_Support]
          [Data\_Privacy]
        ]
      ]
      [Core\_Ethical\_Concepts
        [rdf:type
          [Class]
        ]
        [hasSubConcept
          [Fairness
          ]
          [Accountability
          ]
          [Transparency
          ]
        ]
      ]
    ]
  ]
  \end{forest}
  \caption{An example of ontological block for ethical AI in RDF (Resource Description Framework) form}
  \label{fig:EthicalAIOntology}
\end{figure*}


\begin{figure}[ht!]
    \centering
    \includegraphics[width=0.55\linewidth]{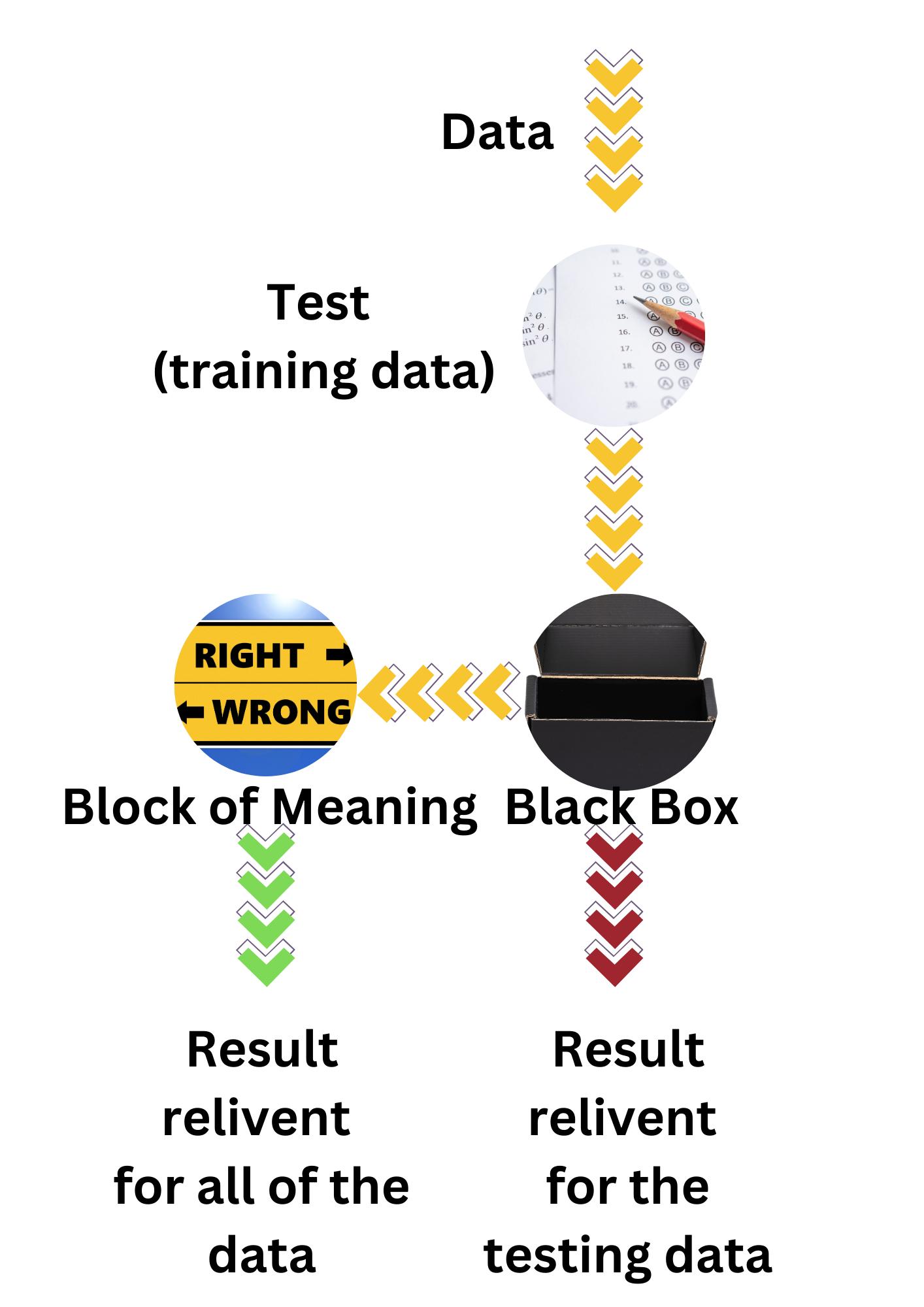}       
    \caption{Sketch: (a) Abstract view of ontological blocks processing for ethical frameworks.}
    \label{fig:EthicalAIOntologyRDF}
\end{figure}


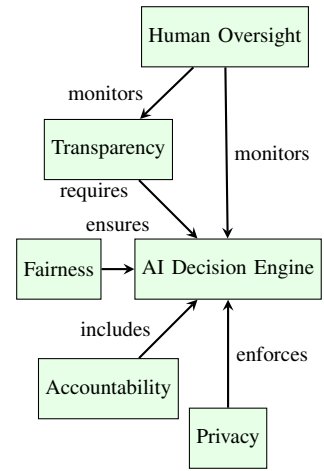
\begin{figure}[ht!]
    \centering
    \footnotesize
    \begin{tikzpicture}[node distance=2.25cm]

    \node (ai) [rdfnode] {AI Decision Engine};
    \node (transparency) [rdfnode, above left of=ai] {Transparency};
    \node (fairness) [rdfnode, left of=ai] {Fairness};
    \node (accountability) [rdfnode, below left of=ai] {Accountability};
    \node (privacy) [rdfnode, below of=ai] {Privacy};

    \draw [arrow] (transparency) -- (ai) node[midway, above] {\hspace{-7em}requires};
    \draw [arrow] (fairness) -- (ai) node[midway, above, yshift=0.4cm] {ensures};
    \draw [arrow] (accountability) -- (ai) node[midway, above, xshift=-0.7cm, yshift=-0.2cm] {includes};
    \draw [arrow] (privacy) -- (ai) node[midway, left, xshift=1.17cm] {enforces};

    \node (human) [rdfnode, above of=transparency, xshift=1.55cm, yshift=-0.75cm] {Human Oversight};
    \draw [arrow] (human) -- (transparency) node[midway, right] {\hspace{-6em}monitors};
    \draw [arrow] (human) -- (ai) node[midway, right] {monitors};

    \end{tikzpicture}
    \caption{Sketch: (b) Representation of ontological blocks for responsible AI. Each node represents a principle, and the relationships define their role in the AI system.}
    \label{fig:rdf_representation}
  
\end{figure}


\subsection{Methodology}
This article uses descriptive research and qualitative observation to explore ethical AI pathways. Key variables include:
\textit{Ethical outcomes} – Benchmarks set by experts in ethics, AI, and policy;
\textit{Verification} – Methods to automate assessments against those benchmarks;
\textit{Scalability} – Applicability across varied AI systems and domains.

Rather than conducting empirical simulations, the study synthesizes literature, expert insights, and theoretical models to propose actionable frameworks.


\subsection{Flexibility Requirements for an Ethical AI}
Integrating ethical AI into legally binding decisions requires acknowledging that legal standards vary significantly by context. Ethical and legal thresholds are not fixed; they shift based on specific circumstances and competing interests. Therefore, ethical AI systems must be adaptable, aligning with legal norms while upholding fairness and accountability.


For example, in serious criminal investigations like murder, authorities may justifiably relax privacy protections to prioritize truth-seeking over individual rights. In contrast, financial transactions—such as a custodian bank managing client funds—demand stricter ethical and legal adherence, emphasizing fiduciary duty over operational flexibility. These scenarios highlight the need for dynamic ethical standards that adjust to context.

This variability challenges the one-size-fits-all ethics frameworks found in much of the AI literature, which often overlook the contextual nature of legal and ethical reasoning. An AI system cannot apply identical principles to a police investigation and a banking transaction, given their differing priorities and risks.

We propose a flexible, modular framework based on \textbf{ontological blocks of meaning}—discrete ethical principles that can be layered or combined. This allows AI systems to dynamically tailor their ethical reasoning to specific legal domains. For instance, an AI analyzing financial data might prioritize fiduciary responsibility, while one used in criminal justice may focus on due process. Such a system ensures ethical adaptability, legal alignment, and transparent oversight.


\subsection{Ontological Blocks of Meaning}
Ontology—the philosophical study of definitions—offers a foundation for aligning human ethical concepts with AI systems. Just as the concept of \textit{light} evolved from Newtonian particles to quantum duality, terms like \textit{responsibility} or \textit{ownership} must be redefined for AI interpretation while preserving their ethical core.

In AI ethics, this means translating abstract ethical terms—\textit{right}, \textit{wrong}, \textit{ownership}, \textit{responsibility}—into constructs that AI can process. These must be both philosophically robust and operationally viable.
Ontological blocks are modular, structured representations of ethical principles that can be integrated into AI operations. Creating them involves:
Translating ethical concepts into machine-readable constructs; Embedding them into modular, systematic representations; Preserving ethical meaning while ensuring scalability and precision.

This ensures AI systems align with human values while providing a functional, auditable ethical framework.

Practically, creating ontological blocks involves developing a structured system to categorize ethical principles, values, and guidelines. Drawing from ontological practices in fields like oncology—e.g., the Cancer Care Treatment Outcome Ontology \cite{lin2018cctoo} and the NCI Thesaurus \cite{kumar2005oncology, ceusters2005ncit}—this process starts by defining core ethical dimensions such as fairness, accountability, and transparency, and mapping them to relevant AI domains.

Collaboration with domain experts ensures term accuracy and contextual relevance. Ethical AI frameworks such as those by Prem \cite{prem2023ethical} support alignment with current standards. Recent models like the Ontology for Ethical AI Principles (AIPO) use vocabularies such as Dublin Core, SKOS, FOAF, and DCAT2 to generate dynamic knowledge graphs, enabling semantic querying and systematic analysis \cite{harrison2021ontology}. These approaches support transparency, consistency, and accountability in ethical AI development.

\subsection{Designing Ontological Blocks for Algorithmic Use}

Ontological blocks translate ethical concepts into structured, quantifiable forms. For instance, defining \textit{“stealing is bad”} involves breaking it down into core concepts like \textit{property} and \textit{ownership}.

\textit{Example:} A dataset element (e.g., a car) represents property, while ownership (e.g., ``Tom owns the car'') provides context. \textit{Generalization:} Instead of rule-based examples, AI evaluates principles of ownership, avoiding reliance on predefined scenarios. \textit{Incomplete data:} AI may infer missing elements through cross-referencing. If cross-references are weak, human oversight is required. \textit{Short format:} The point of the ontological block is that it would refer a single concept (e.g. ownership), therefore more complex ethical questions would be described as a sum of multiple ontological blocks. 

\subsection{Common Structure of Ontological Blocks of Meaning}

A major challenge in ethical AI is ensuring ontological blocks follow a common structure to support uniform testing and consistent term relationships. The proposed structure is simple: a primary concept paired with a binary ethical qualifier (e.g., \textit{good} or \textit{bad}).

For example, in the block \textit{“stealing is bad”}, \textit{stealing} is defined as the unauthorized transfer of ownership outside quantifiable pathways like trade, gift, bequest, or taxation. This format supports clear, consistent, and transparent ethical evaluations.

Block variables are selected by designers—researchers, practitioners, or stakeholders—ensuring both expert insight and public auditability. For complex concepts (e.g., \textit{medical emergency}), multiple simpler blocks can be combined. This modular structure supports consistency and scalability across domains.
 
These are the foundational principles that could guide visualizing ethical AI ontological block. They include:

\textbf{Fairness:} Prevents bias and ensures equitable treatment.
\textbf{Accountability:} Assigns responsibility for AI outcomes.
\textbf{Transparency:} Promotes understandable, open AI processes
\textbf{AI Application Areas:} Domains where ethical principles are applied.
\textbf{Ethical Frameworks:} Structured sets of principles guiding ethical AI development. 

Together, these components provide a robust structure for designing ethical AI systems that are principled, transparent, and aligned with current standards.

\subsection{Pros and Cons of the Proposed Framework}

\subsubsection{Advantages}
Independence from training data, enabling robust ethical evaluations. Flexibility to combine blocks across fields, such as medicine or law. Auditability of standardized blocks, improving transparency and accountability.

\subsubsection{Drawbacks}
Labor-intensive creation of non-contradictory, quantifiable blocks. Reliance on large language models (LLMs) or external tools to infer missing relationships, introducing potential biases. Additional processing power required for integration and maintenance.

Despite challenges, the framework’s flexibility, independence, and auditability make it a promising foundation for building ethically aligned AI systems.

\subsection{A Use-Case: AI-Powered CRM and Ontological Profiling for Investor Protection}

\textbf{Overview:} Brokerage firms are increasingly using AI-powered voice assistants to enhance investor classification and risk profiling. Moving beyond static binary systems, these tools enable more precise, ethical, and adaptive assessments of investor risk tolerance, protecting retail clients from unsuitable high-risk products \cite{egbuhuzor2025ai}.

\subsubsection{The Challenge}

Regulations require firms to distinguish between professional and retail investors to prevent the latter from accessing high-risk instruments like synthetic options. However, current assessments often overlook real-world risk tolerance, especially under stress, leading to potential harm and regulatory noncompliance.

\subsubsection{The Innovation}

This use case introduces \textit{ontological blocks of meaning} for product classification. A financial product might trigger a ``Riskier'' block based on:

- \textbf{Financial risk} – Potential economic loss.

- \textbf{Psychological risk} – Emotional reaction to financial stress.

This enables real-time, context-aware, and ethically grounded risk assessment.

\subsubsection{How It Works}

AI voice assistants engage clients with adaptive questions (e.g., ``How did you react to your last investment loss?''). A response like ``Not well'' may indicate emotional sensitivity. Using NLP and probabilistic modeling, the system estimates a behavioral risk profile.

If there's a high probability (e.g., 60\%) of emotional vulnerability, a corresponding ontological block is triggered, ethically restricting access to high-risk products. This creates a personalized, behavior-informed investor classification framework.


\subsection{Discussion}

The evaluation shows that combining ontological blocks with FAIR principles offers a strong foundation for ethical AI monitoring. Key insights include:
\begin{itemize}
    \item The unified monitoring system effectively identifies ethical risks and supports compliance (e.g., EU AI Act).
    \item Ontological blocks provide structured, modular, and interpretable ethical encoding.
    \item FAIR principles enhance discoverability, usability, and scalability.
\end{itemize}

Remaining challenges include the manual effort required to build blocks and the reliance on probabilistic reasoning with incomplete data. Future work should focus on automating block creation and improving data integration.

\section{Conclusion and Recommendation}
\label{sec:ConclusionAndRecommendation}

This framework introduces ontological blocks as standardized, expert-designed tools for ethical AI assessment. These modular units enable transparent evaluations across diverse ethical contexts.

Key strengths include independence from training data and flexibility across domains. However, challenges remain in automating block creation and managing dependencies on probabilistic models.

By integrating with FAIR principles, the framework supports ethical, transparent, and accountable AI systems—laying the groundwork for responsible AI development across sectors.

\section*{Acknowledgment}
The authors sincerely thank theGesellschaft für wissenschaftliche Datenverarbeitung mbH Göttingen (GWDG, Germany) for their valuable contributions. We are especially grateful for support from NHR\footnote{\url{http://www.nhr-verein.de/en/ourpartners}}, essential for presenting this work, and appreciate the constructive feedback from our peers.
This research was funded by the EU KISSKI Project under grant number 01|S22093A (Förderkennzeichen).

\bibliographystyle{IEEEtran}
\bibliography{references}

\end{document}